# Role of Simplicity in Creative Behaviour: The Case of the Poietic Generator

**Antoine Saillenfest\*, Jean-Louis Dessalles\*, Olivier Auber\*\***

\* Telecom ParisTech, LTCI, Université Paris-Saclay, Paris, France
\*\* ECCO, Free University of Brussels (VUB) & The Global Brain Institute, Brussels, Belgium
{antoine.saillenfest, jean-louis.dessalles}@ telecom-paristech.fr - olivier.auber@vub.ac.be

## Abstract

We propose to apply Simplicity Theory (ST) to model interest in creative situations. ST has been designed to describe and predict interest in communication. Here we use ST to derive a decision rule that we apply to a simplified version of a creative game, the Poietic Generator. The decision rule produces what can be regarded as an elementary form of creativity. This study is meant as a proof of principle. It suggests that some creative actions may be motivated by the search for unexpected simplicity.

## Introduction

Can human creativity be captured by equations or algorithms? The idea seems contradictory. Most creative acts are by essence unexpected and cannot be predicted. But if unexpectedness is the hallmark of creativity, couldn't we use it as a proxy? Our hypothesis is that *creative processes should maximize unexpectedness*.

To test the hypothesis, we considered a situation that is sufficiently constrained to offer a limited range of possibilities, but that is still rich enough to give rise to creative behaviour. We used a simplified version of the "Poietic Generator" for that purpose. Our point is to offer a proof of principle by showing that a program implementing the principle of maximum unexpectedness may mimic creative behaviour.

This study relies on a formal definition of unexpectedness provided by Simplicity Theory. To be unexpected, creative acts must generate some complexity drop for an observer. This principle proves sufficient, in the constrained situation of the Poietic Generator, to produce non trivial patterns of actions that can be regarded as creative.

In what follows, we briefly introduce Simplicity Theory and the notion of unexpectedness. We then describe the simplified version of the Poietic Generator that we have been using for our experiments. We then explain how we implemented the principle of maximum unexpectedness in that game and show our results. Lastly, we discuss how this basic form of creativity can be used as a basis to analyse more complex creative behaviour.

## Unexpectedness and Simplicity

Our hypothesis is that to appear creative, actions should involve unexpected aspects (Bonnardel, 2006; Maher, 2010). In some situations such as the one analysed here, the set of available actions is so limited that a good way of achieving creativity consists in adopting the following principle:

*Principle of maximum unexpectedness in creativity:*

> Select actions
> that will maximize
> unexpectedness.

There are few formal definitions of surprise or unexpectedness. Schmidhuber distinguishes between (un)predictability, unexpectedness, surprise and interestingness (Schmidhuber 1997a; 1997b; 2003; 2009). For him, unpredictability implies unexpectedness, but unexpectedness does not imply surprise, which is defined with reference to expectations (Schmidhuber, 2003). He also defines interest as the time-derivative of the best compression an observer can achieve from the situation (Schmidhuber 2009). This means that interest is raised when the observer is making more sense of the current situation.

The framework of Simplicity Theory[1] (ST) also makes use of a difference in complexity. ST was introduced to explain why some events are unexpected and newsworthy (Dessalles, 2006; 2008). *Unexpectedness* is defined as the difference between expected complexity and observed complexity.

$$U = C_{exp} - C_{obs}. \qquad (1)$$

The term 'complexity', also known as Kolmogorov complexity, refers to its theoretical definition, namely the size of the shortest summary. We do not consider objective complexity, which is not computable (Li & Vitányi, 1994), but a resource-bounded version of it (Buhrman *et al.*, 2002). ST introduces a difference between $C_{exp}$ and $C_{obs}$. The former is generally assessed through the complexity of a *causal* pro-cedure, whereas the latter is free from this constraint and matches the usual definition of (resource-bounded) complexity. This difference between generation and observation is parallel (though not identical) to the difference between 'generation' and 'distinction' (Buhrman *et al.*, 2002). ST designates causal complexity by $C_w$. Since $C_{obs}$ corresponds to a minimal description, it can be noted $C_d$. The unexpectedness of an observed event can be rewritten as:

$$U = C_w - C_d. \qquad (2)$$

---
[1] See www.simplicitytheory.science.

This definition explains why the content of a blank page is not unexpected ($C_w = C_d = 0$) and why a random binary string of size $n$ is not unexpected either ($C_w = C_d = 2^n$). A remarkable lottery draw like 1-2-3-4-5-6, on the other hand, would appear extremely unexpected: all draws have same causal complexity, as they require the generation of 6 numbers ($C_w \approx 20$ bits) ; most of them require the enumeration of 6 numbers as well to be unambiguously determined ($C_d \approx 20$ bits), but not the consecutive draw which can be described with much less ($C_d \leq 3$) (Dessalles, 2006).

In toppling domino challenges, flicking one single domino leads millions of them to fall down. The global result is spectacular; particular moments revealing a well-known image or triggering some mechanical device such as a tiny catapult are spectacular as well. Does our definition of unexpectedness account for such effects? The intended results are certainly chosen to be mostly simple (*i.e.* they require a small amount of information to be described): all dominoes down, a well-known image revealed, a world record to break. But the highly complex causality leading to these events plays a crucial role as well. Even when the process is going on, one can measure the number of failure opportunities (any domino may fail to fall down) that make $C_w$ quite huge. Unexpectedness, and therefore interest, comes from the contrast between both complexity values.

We tested the idea that creative acts appear all the more creative as observers are able to see them as unexpected. In other words, the end result of a creative act must be both simple and seemingly hard to reach.

## The Poietic generator

To test the role of unexpectedness on creativity, we had to find a situation in which the machine may explore a limited, but still rich, gamut of actions. We also wanted to stay close to a situation of artistic creativity, where no predefined task is to be fulfilled. The Poietic Generator, created in 1986 by Olivier Auber, offered us an ideal framework.

The Poietic Generator (PG) is a game with no rule. All players see the same matrix, displayed on their screen, but they control only one cell of the matrix. In the real game (which is ongoing: anyone can connect to http://poietic-generator.net/ and play), players have a rich control of their portion of the screen, in which they can draw coloured shapes. In the absence of any instructions, players start creating what might look like a random pattern from a distance. But the collective tends to self-organize somewhat, with structures emerging from time to time, either locally or globally, as shown in Figure 1 (see also animated recordings at http://tinyurl.com/pgen1).

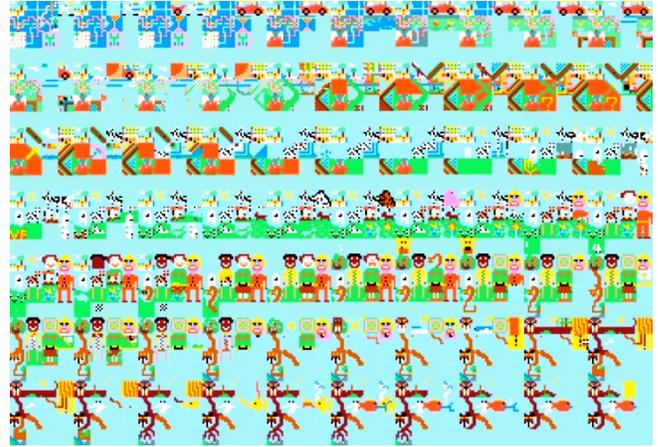

*Figure 1: Successive states in a PG session observed in 1996 at Telecom ParisTech (9 participants, ~30 min).*

We decided to study how programs would behave on the Poietic Generator if they followed the "principle of maximum unexpectedness." The point is to see whether the machine appears to be creative and in what sense. This poses several challenges. First, we must define what the machine observes and how it computes generation and description complexity. Second, we must define what we would consider as 'being creative'. And third, we should compare the productions of the machine with what humans do.

To make the three challenges manageable, we had to simplify the game significantly. In our simplified version of the Poietic Generator (SPG), each cell of the matrix consists in one pixel. Each player, as a result, can only control one among $K$ colours. Moreover, all players are instantiations of the same decision rule. Even this way, the SPG remains rich enough to offer the opportunity of being creative. It is not easy to predict what will happen, and it is not easy to tell in advance what creative actions would be. However, it might be easier to tell after the fact that reaching such or such state was (somewhat) creative.

## Coding representations

Any artificial creative device must rely on a model of aesthetic preference. In our approach, unexpectedness provides such a hierarchy. The computation of $U$, however, presupposes a cognitive model from which complexity can be computed, for instance a neo-Gestaltist theory in which simple patterns are group invariant (Leyton, 2006). To keep things simple, we decided to use a set of pre-computed simple patterns that we call *basic patterns*. This is, of course, only acceptable for the SPG, and even for the SPG, for small matrix sizes.

Figure 2 shows a rudimentary set of monochrome basic patterns that we used to evaluate the SPG. Our implementation of the SPG, however, accepts coloured basic patterns. For instance, the definition of a trichromatic pattern relies on three colours ($c_1$, $c_2$, $c_3$) that are not instantiated. The distance $H(s_c, p)$ from a given state $s_c$ to a pattern $p$ counts all differing cells between $s_c$ and $p$ for each possible colour instantiation of the pattern and keeps the minimum value. In the monochromatic case, computing $H(s_c, p)$ amounts to taking the minimum between two Hamming distances.

*Figure 2: Example of basic 5×5 monochrome pattern set.*

The originality of our model is not only that it is based on the notion of complexity, but also to distinguish generation from description. Generation complexity $C_w$ depends on players' actions. In the SPG, the minimal causal history leading from a reference state $s_r$ to a target state $s_t$ consists in indicating the location of each differing cell and how it should be switched. In a SPG of size $n \times n$, one needs $2 \times \log_2(n)$ bits to designates a cell in the matrix. The number of differing cells between $s_r$ and $s_t$ is $H(s_r, s_t)$. For each differing cell, one needs to indicate the target colour. If $K$ is the number of available colours, we need $\log_2(K)$ bits to designate the correct colour among the $K$ alternative decisions. The complexity of generating the transition $s_r \rightarrow s_t$ can be written as the minimum amount of information needed to transform $s_r$ into $s_t$.

$$C_w(s_r \rightarrow s_t) = H(s_r, s_t) \times (2\log_2(n) + \log_2(K)). \quad (3)$$

The decision rule consists in searching a maximally unexpected pattern. If we write $\alpha = 2\log_2(n) + \log_2(K)$, equation (2) now reads:

$$U(s_t) = \alpha H(s_r, s_t) - C_d(s_t). \quad (4)$$

Equation (4) provides a hierarchy of attractiveness for the set of target patterns $\{s_t\}$. At the beginning of the game, $s_r$ is set to the initial configuration of the grid. If the initial state is random, then $H(s_r, s_t)$ has roughly the same value for all target states $s_t$. As a consequence, the most attractive targets are the simplest ones: all-white and all-black.

The complexity of reaching targets may however obliterate their attractiveness. ST takes this complexity into account to determine how much actions and targets are wanted (Saillenfest & Dessalles, 2014). We transpose this notion to the SPG by defining the *desirability* of a given target $s_t$ seen from the current state $s_c$:

$$D(s_c, s_t) = U(s_t) - \alpha H(s_c, s_t). \quad (5)$$

The term $\alpha H(s_c, s_t)$ represents the complexity of generating $s_t$ from the current state $s_c$. We can see that the most attractive states are not necessarily the most desirable ones. If we put (4) and (5) together, we get:

$$D(s_c, s_t) = \alpha H(s_r, s_t) - C_d(s_t) - \alpha H(s_c, s_t). \quad (6)$$

There is a trade-off between three terms: the difficulty of reaching the target from the reference state, its overall simplicity and the easiness of reaching it from the current state.

The distinction between the reference state and the current state is crucial here. Players are trying to produce an event that will appear unexpected to a hypothetical audience. The audience may be the community of players currently acting on the grid. It may also be anyone connected to the game just for watching in the case of the real Poietic Generator. For this audience, a pattern will constitute an event if it is unexpected as compared to the initial state, or later to subsequent reference states. In the case of the toppling dominoes, the final event: all dominoes having fallen down, is only unexpected in comparison with the initial state.

Our artificial SPG players base their strategy on a similar comparison (equation (4)). When selecting an action to perform, however, they measure the distance from the current state to tentative goals (equation (5)). They begin by selecting a mostly desirable goal $s_t^\circ$.

$$s_t^\circ = \operatorname{argmax}(D(s_c, s_t)). \quad (7)$$

They change their colour only if it increases desirability, which amounts to saying that $H(s_c, s_t) > H(s_c', s_t)$, where $s_c'$ is the state resulting from their changing colour. Using this strategy, the system is expected to converge on a simple state, not necessarily a simplest one. Once such a target is reached, the reference is set to that new state for all players: $s_r = s_t^\circ$. Due to this change, $s_t^\circ$ is no longer desirable, as it is no longer unexpected. The community of players starts looking for another goal that it may then reach, and so on. The emerging result is that the SPG will visit various simple states in this way. This travel through the state space in search for simplicity generates a basic form of creativity.

## Implementing the SPG

The SPG is initialized as an $n \times n$ matrix, where each cell is set to white or, alternatively, assigned a random colour.

Each agent controls one cell of the matrix. It stores the initial global state of the game as its reference state. At each time step, one among the $n^2$ agents is randomly selected to play. This agent decides either to change the colour of its cell or do nothing, depending on the decision procedure described below. If the system reaches a state $s_t$ that is maximally desired according to (6), then $s_t$ becomes the new reference state for all agents.

## Evaluating desirability

When an agent is selected to play, it has to decide whether to change colour or not. To do so, it evaluates the desirability of reachable target states using (6). Figure 3 illustrates how values of $\alpha H(s_c, s_t)$ are compared for different targets $s_t$. When the target is a multicoloured pattern $p$, e.g. a trichromatic pattern with variable colours $(c_1, c_2, c_3)$, we add up the three Hamming distances computed only over pixels that are $c_i$-coloured in the pattern. The same computation is done over all instantiations of $(c_1, c_2, c_3)$; the smallest result is taken as $H(s_c, p)$ and instantiates $p$.

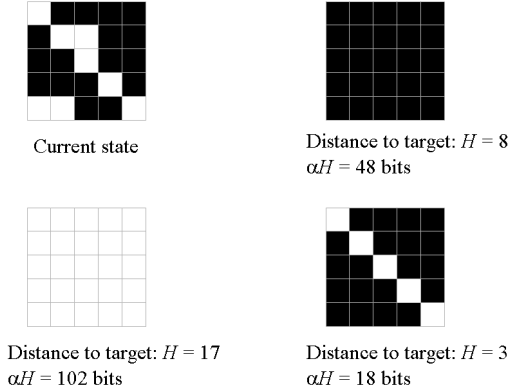

Figure 3: Distance to 3 basic states in a 5×5 monochrome SPG.

It is reasonable to consider that some target states are "too far" from the current state of the SPG to be considered by an agent as candidate target states. We introduce the notion of *horizon*, which is the value of $H(s_c, s_t)$ beyond which agents do not evaluate potential target states. In other words, a state $s_t$ is a candidate target state only if:

$$H(s_c, s_t) \leq horizon. \quad (8)$$

To compute desirability according to (6), we need to estimate $C_d(s_t)$ for any basic state $s_t$ (figure 2). Details are not crucial here, as long as the computation provides a reasonable hierarchy of forms. The code we chose is based on the following tuple that describes any basic state:

(*colours*, *shape*, *configuration*)

- *colours* is a tuple defining the pattern's colours. For a $q$-chromatic pattern, $\log_2(K!/(K{-}q)!)$ bits are sufficient to determine the *colours* tuple unambiguously.

- We need at most $\log_2(nshape)$ bits to discriminate the pattern's *shape* among the *nshape* basic shapes (in the example of figure 3, only three shapes are considered: diagonals, lines and triangles).
- We need at most $\log_2(nconfig)$ bits to determine one configuration among the *nconfig* configurations that correspond to the same shape. For example, in figure 2, there are 10 possible configurations for the shape 'line'.

We approximate the description complexity of a basic state $s_t$ using these upper values:

$$C_d(s_t) = C_d(colours) + C_d(shape) + C_d(configuration \mid shape).$$

We get (logarithms are approximated by their upper integer value):

$$C_d(s_t) = \log_2(K!/(K{-}q)!) + \log_2(nshape) + \log_2(nconfig). \quad (9)$$

Note that *nshape* and *nconfig* depend on $s_t$. Formula (9) can be used to rank basic states by simplicity (see Table 1).

*Table 1: Description complexity of basic states in a monochrome 5×5 matrix with 3 possible shapes (diagonals, lines, triangles).*

| Basic SPG state | (colours, shape, configuration) |
|---|---|
| *colours*: (white, black) or (black, white) *shape*: None (■ or ▦) | Example of code: (all-black) `[0, _, _]` Length: 1 + 0 + 0 = 1 bit |
| *colours*: (white, black) or (black, white) *shape*: Diagonal (ex: ▨ , ▨ ) | Example of code: (white rising diagonal) `[0, 00, 0]` Length: 1 + 2 + 1 = 4 bits |
| *Colours*: (white, black) or (black, white) *shape*: Triangle (ex: ◣ , ◤ ) | Example of code: (white upper right triangle) `[0, 01, 01]` Length: 1 + 2 + 2 = 5 bits |
| *Colours*: (white, black) or (black, white) *shape*: Line (ex: ▤ , ▤ ) | Example of code: (white second horizontal line) `[0, 10, 0101]` Length: 1 + 2 + 4 = 7 bits |

## Decision procedure

Once the desirability of candidate target states has been computed by agents, two things may happen: either one candidate state (ore more) is maximally desirable, or none is desirable. When at least one state is maximally desirable, it becomes the agents' current target. Agents change their colour only if it brings them closer to the target for the $H(s_c, s_t)$ distance. Otherwise, agents perform no action.

At the beginning of the game, desirability is the same for all candidate states $s_t$ (equation (6)): $D(s_c, s_t) = D(s_r, s_t) = -C_d(s_t)$. The same holds when a reference state has been reached and the new reference is taken to be $s_r = s_c$. In these situations, none of the possible target states is desirable ($D(s_r, s_t) < 0$). A rational agent should not act in the absence of goal. However, such an attitude would be counter-productive in a creative context. In the case of the SPG, the game would freeze, since all agents have the same reference state and the same horizon. We programmed agents to change colour with a certain probability in the absence of desirable state.

## Results

Though we implemented SPG for an arbitrary number of colours, we evaluated it only for $K = 2$ (black and white). Our results vary somewhat depending on the value of *horizon*. Figure 4 displays the fraction of the time during which agents are able to find desirable states, as a function of *horizon*. For small values of *horizon*, agents do not "see" any target state and no state is ever desired. At the other end, with a high value of *horizon*, agents can see desired states all the time and are always in goal-oriented search. Moreover, the distribution of desired states shows that the most desired states correspond most of the time to the simplest ones (*i.e.* the two plain states). For some intermediate value of *horizon*, agents target a broader range of states.

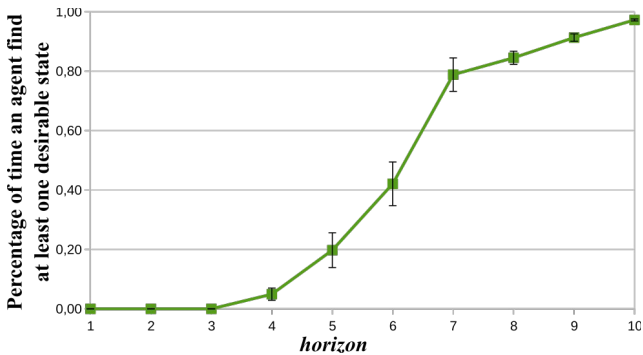

*Figure 4: Fraction of time playing agents find at least one state desirable (matrix 5x5).*

In the results presented here, parameters are fixed at the following values:

- size of the matrix: 5×5, two colours, horizon: 7;
- when no state is desirable, agents change colour with probability 0.5.

When the program runs, the state of the SPG changes continuously (see http://spg.simplicitytheory.science). From time to time, it reaches a low-complexity state. Figure 5 shows how the complexity of the SPG matrix evolves through time. In this figure, the complexity of the current state $s_c$ is computed in reference to the closest basic state $p$, by evaluating $\min_p(H(s_c, p) + C_d(p))$. We can observe its "oscillations" as it visits basic states (figure 2) and then moves away from them.

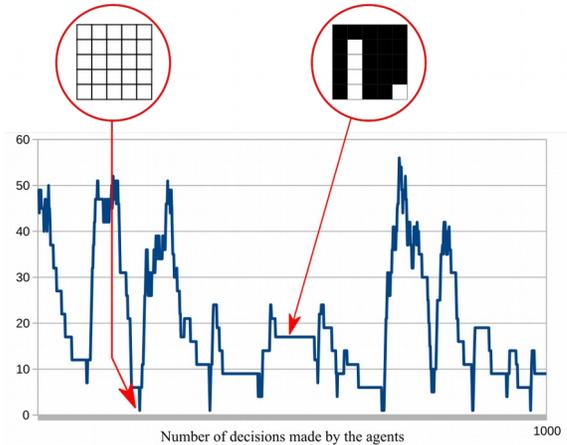

*Figure 5: Evolution of the description complexity of the SPG matrix over the first 1000 individual decisions.*

Figure 6 corresponds to the same run. It shows the 100 first target states that were successively reached. We can see that the system visits almost all the "basic states" described in figure 2. A simple analysis using periodograms did not reveal any regularity in this sequence.

Note that transitions are not totally random. Figure 6 reveals that when a plain state has been reached, the next basic step is likely to be a diagonal (among the 27 transitions from a plain state to another basic state, 20 lead to a diagonal state with same background). This observation makes sense. Seen from a plain state, diagonals with same background colour lie at Hamming distance 5, which is smaller than *horizon*. For most agents, changing colour would not bring them any closer. Nine of them, located on the diagonals, can get closer by one unit to a diagonal pattern, which becomes more desirable by $\alpha \approx 6$ bits (formula (6)). Since its description complexity amounts to 4 bits (Table 1), its desirability is now 2 bits. If one of those 9 agents is by chance next to play, it will change colour. The diagonal then becomes desirable to all agents. As a result, the probability that the next target will be a diagonal when starting from a plain state must be larger than $9/25 = 0.36$. We measured 0.44. Table 2 shows shape to shape transition frequencies computed after a single run of the program (the number of observed transitions from a given shape to another shape is indicated under the reference shape).

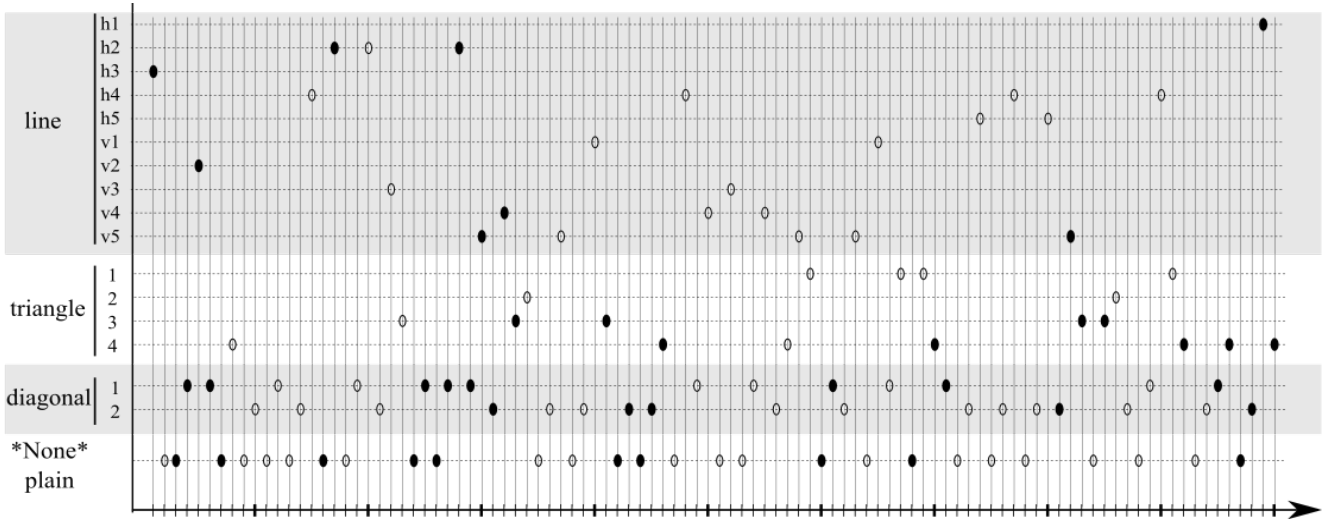

*Figure 6: Visited shapes in chronological order. Dot colours (black or white) correspond to shape background.*

*Table 2: Frequency of transitions between shapes measured in a 5×5 matrix for the shapes of Table 1 during a single run of the program (bc=same background colour; 1-bc = opposite).*

| | | Next shape | | | |
|---|---|---|---|---|---|
| | | Plain | Diagonal | Triangle | Line |
| **Shape taken as reference** | Plain (1162 transitions) | **0.1** bc: 0 1−bc: 0.1 | **0.44** bc: 0.44 1−bc: 0 | **0.23** bc: 0.02 1−bc: 0.21 | **0.24** bc: 0.24 1−bc: 0 |
| | Diagonal (764 transitions) | **0.26** bc: 0.23 1−bc: 0.03 | **0** bc: 0 1−bc: 0 | **0.12** bc: 0.06 1−bc: 0.06 | **0.62** bc: 0.6 1−bc: 0.02 |
| | Triangle (999 transitions) | **0.50** bc: 0.16 1−bc: 0.34 | **0.07** bc: 0.07 1−bc: 0 | **0.26** bc: 0 1−bc: 0.26 | **0.17** bc: 0.17 1−bc: 0 |
| | Line (906 transitions) | **0.35** bc: 0.22 1−bc: 0.13 | **0.22** bc: 0.21 1−bc: 0.02 | **0.43** bc: 0.23 1−bc: 0.20 | **0** bc: 0 1−bc: 0 |

## Discussion

We showed that a simple strategy, the "principle of maximum unexpectedness", leads to seemingly creative actions. From the definition of unexpectedness as complexity drop between generation and description (formula (2)), we derived the strategy of maximum of desirability (rules (6) and (7)) that continually looks for simple patterns within reach. This strategy, together with the notions of reference state and of horizon, is sufficient to generate interesting behaviour in the SPG. When several instances of the strategy play together the SPG, we observe an emerging behaviour which consists in visiting simple patterns in an unpredictable way (see examples at http://spg.simplicitytheory.science).

The walk through simple patterns is the best (in the sense of most creative) we could get in this simple implementation, at least from a theoretical point of view. Any other emerging result among what the system could have achieved (fixed point, periodic behaviour, random walk) would have been less creative. The program mimics the following features of creativity.

- Search for unexpected simple patterns,
- Co-existence of goal-free and goal-oriented actions,
- On-going goal change,
- Fresh start when a goal is achieved.

Unexpectedly simple patterns are essential to most forms of artistic creativity. One extreme example is offered by the "White on White" painting exhibited by Kazimir Malevich in 1918. Note that further instances of so-called monochrome paintings (*i.e.* uniformly coloured surfaces) can be felt as less creative than the very first one, as they are more complex (more information is needed to discriminate them from each other). More generally, any hidden structure discovered by the observer in a painting makes it more interesting (Leyton, 2006). According to Leyton, the more circles and ellipses our eye can see in Picasso's "Les demoiselles d'Avignon", the more beautiful the painting appears. This makes sense within ST's framework: hidden structure means unexpected simplicity. Any structural component contributes to simpler description, as previously independent components can now be summarized by the structure.

In contrast to routine engineering activity which is goal-driven, some artistic activities are carried out in the absence of definite goal and they are able to invent their own goals on the fly. This is what our implementation of the SPG does, despite its elementary character. This absence of pre-definite goal is perceived by observers. As predicted by (1), the interplay of seemingly random actions and oppor-tunistic goal generation produces a series of complexity drops that may trigger feelings of beauty (Schmidhuber, 2009). Note, however, that the originality of ST is to define compression through (2) by making a distinction between generation and description.

Our implementation of the SPG has elementary self-observation capabilities. It knows when an interesting creation has been reached. Our program stops for a while when it gets to a target configuration. These moments correspond to local complexity minima. Then, by setting the reference to the former goal state, the system is able to escape from it. As equation (6) shows, the state is no longer desirable once it has become the new reference. The system automatically hunts for new simple states to spot and reach.

When no desirable state is within reach, the decision rule expressed in (6) and (7) does not apply. In such situations of goal-free action, human individuals tend to perform actions anyway, though in a biased manner (Auriol, 1999). Our simulated players are not biased and switch their state randomly. This aspect of our implementation is not governed by any principle and could be improved in more elaborate versions of the SPG.

The desirability of target states expressed by (6) is interesting because it includes two versions of generation complexity $C_w$. We can write it in a more general form.

$$D(s_c, s_t) = C_w(s_r \rightarrow s_t) - C_d(s_t) - C_w(s_c \rightarrow s_t). \quad (10)$$

The latter term, $C_w(s_c \rightarrow s_t)$, represents a low complexity value that was not anticipated when the system started from the reference state. A target $s_t$ is desirable if $C_w(s_c \rightarrow s_t) << C_w(s_r \rightarrow s_t)$, which means that $s_t$ is significantly easier to reach than anticipated. The same phenomenon holds in other forms of creative productions, such as fiction writing. Interest in a narrative may be aroused when some surprising event occurs, but then is perceived as making sense after all, because of some hidden line of reasoning (Saillenfest & Dessalles, 2014; Saillenfest, 2015). Rule (10) offers a similar kind of surprise when the system comes close to a simple state that was initially considered out of reach, but now appears to be just a few moves away.

## Limits and perspectives

Our experiment has obvious limits. It is a proof of concept that does not aim at giving an illusion of genuine creativity. We are indeed quite far from what a human observer would regard as truly creative.

Human players in the true PG show significantly more sophisticated behaviour. They may form individual intentions based on their personal history and context; they are able to recognize many shapes and not only geometrical ones: horses, houses or human figures; on the other hand, they have limited patience and attention span. Differences among individual players may lead to paradoxical situations, as when an idle player becomes a stable anchor around which local activity gets organized, and emerges as a local attractor for this activity.

Human players may collectively produce simple emerging patterns such as a uniform area or a checkerboard. Most emerging patters, however, consist in recognizable shapes: a cow, a monster, a sea shore, an air strike or a luncheon on the grass. These patterns may occur in one part of the global image and may be inspired by the news.

Mimicking these human capabilities depends on the system's ability to select and recognize elaborate patterns. If the size of the matrix is increased, the set of basic patterns (lines, triangles…) becomes too sparse for the SPG to see any target within the horizon. The situation would become even more complicated if the set of possible actions is increased to bring the SPG closer to the true PG: many colours, more pixels controlled by each player. Populating the set of simple shapes may solve the sparseness problem. However, the problem of computing the complexity $C_d$ of elaborate shapes is not a trivial task (think of recognizing an air strike and its relation to the news).

Another characteristic behaviour exhibited by human players consists in calling attention to themselves whenever possible. For instance, a player controlling a cell in the middle of a uniform region may be tempted to switch to a locally contrasting colour, as in the yin-yang (Taijitu) symbol. This makes sense within ST. The theory indeed predicts that the complexity drop that drives attention to the individual will be larger when her cell is isolated. Its minimal description will be more concise if it is a contrast with its surroundings. A way to improve our model of creativity would be to include a second complexity drop computation at the individual level, so as to allow artificial players to choose between collective creativity and individual signalling.

SPG can be seen as a first step toward a new class of cellular automata that try to mimic some aspects of human creative behavior. One possible perspective for further developments would be to design artificial agents able to play in real PG games. Human players would be asked if they are able to locate them. This experiment might be seen as a visual version of the Turing test. It could become the basis of an 'open science' initiative to study the specificity of human creative behavior. This open experiment could also help to deal with ethical questions about human-computer entanglement in a manner that would be accessible to all (Auber 2016).

All the above mentioned improvements of the SPG keep fundamental principles derived from Simplicity Theory intact. The decision rule expressed by (10) and (7) would remain essentially the same (except for the pattern distance

(3) which cannot remain based on the Hamming distance if more pixels are controlled by each player). Our little experiment with the SPG was designed to be just sufficient to implement the decision rule. This is why it is relevant to study creativity

## Conclusion

This study was motivated by the observation that simplicity and complexity drop play a crucial role in creativity. We decided to investigate whether Simplicity Theory could make an interesting contribution to the understanding of creative action. ST was developed to offer a formal definition of interest in human spontaneous communication. Quite naturally, we wanted to explore whether interest in creative situations could be governed by similar mechanisms.

The present study is meant as a proof of principle. We proposed a simplified implementation of the Poietic Generator to verify that a straightforward application of ST's principles could lead to interesting behaviour even in a simplistic setting.

The "principle of maximum unexpectedness" (formula (10)) that we derived from ST makes a trade-off between three values: (1) the simplicity of the target, (2) the difficulty to reach it from the reference situation and (3) the easiness to reach it from the current situation. This decision rule is claimed to apply to a wide range of creative situations. We were able to show that these theoretical principles produce non trivial behaviour even in a simplistic situation. Our suggestion is to consider these principles when designing elaborate creative programs.

## Acknowledments

This study was supported by grants from the programme Futur&Ruptures and from the "Chaire Modélisation des Imaginaires, Innovation et Création".